\documentclass[10pt,twocolumn,letterpaper]{article}

\usepackage[pagenumbers]{cvpr} 

\usepackage{algorithm}
\usepackage{algorithmic}

\definecolor{cvprblue}{rgb}{0.21,0.49,0.74}
\usepackage[pagebackref,breaklinks,colorlinks,allcolors=cvprblue]{hyperref}

\title{PALUM: Part-based Attention Learning for Unified Motion Retargeting}

\author{
Siqi Liu$^{1,2}$\thanks{Work was done during Siqi’s internship at Feeling AI.} \quad Maoyu Wang$^{2}$ \quad Bo Dai$^{2,3}$ \quad Cewu Lu$^{1}$
\\[0.5em]
$^{1}$Shanghai Jiao Tong University \quad $^{2}$Feeling AI \quad $^{3}$The University of Hong Kong 
}

\let\oldtwocolumn\twocolumn
\renewcommand\twocolumn[1][]{%
\oldtwocolumn[{#1}{
\begin{center}
    \includegraphics[width=\textwidth]{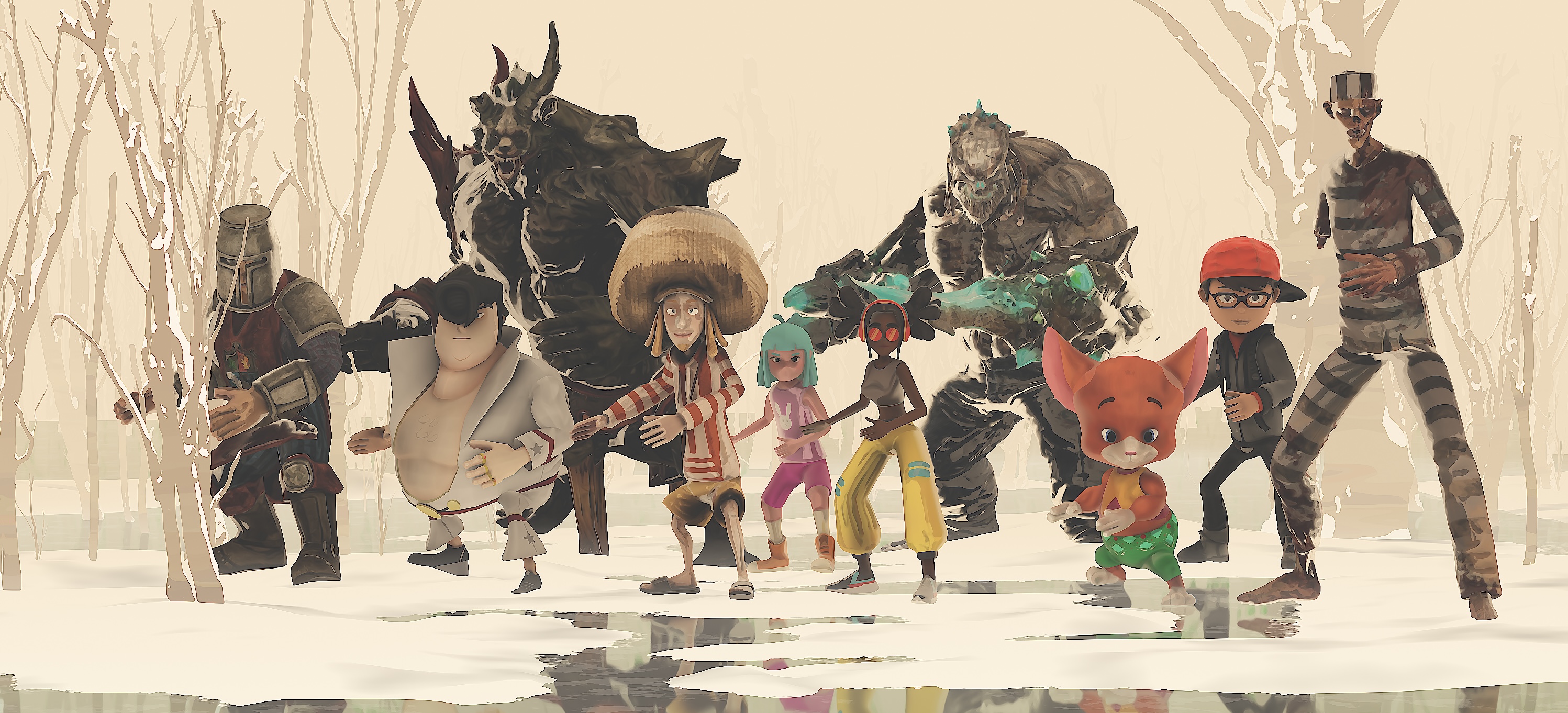}
    \captionsetup{hypcap=false}
    \captionof{figure}{PALUM retargets motion from the source skeleton to skeletons with varying joint numbers and bone proportions. The boy with the red hat is the source character, and others are the target characters.}
\end{center}
}]
}

\begin{document}
\maketitle
\begin{abstract}

Retargeting motion between characters with different skeleton structures is a fundamental challenge in computer animation. When source and target characters have vastly different bone arrangements, maintaining the original motion's semantics and quality becomes increasingly difficult.
We present PALUM, a novel approach that learns common motion representations across diverse skeleton topologies by partitioning joints into semantic body parts and applying attention mechanisms to capture spatio-temporal relationships. Our method transfers motion to target skeletons by leveraging these skeleton-agnostic representations alongside target-specific structural information. To ensure robust learning and preserve motion fidelity, we introduce a cycle consistency mechanism that maintains semantic coherence throughout the retargeting process. Extensive experiments demonstrate superior performance in handling diverse skeletal structures while maintaining motion realism and semantic fidelity, even when generalizing to previously unseen skeleton-motion combinations. We will make our implementation publicly available to support future research.


\end{abstract}    
\section{Introduction}
\label{sec:intro}

Motion retargeting, the process of transferring motion sequences between characters with different skeletal structures, is essential across gaming, virtual reality, film, and robotics industries. As digital content creation scales, studios frequently need to adapt existing motion capture data to characters with vastly different anatomical configurations. Currently, this process relies heavily on manual intervention by skilled animators, creating a time-intensive bottleneck in production pipelines. The need for robust automated cross-skeleton motion retargeting has thus become a fundamental research challenge, promising to accelerate content creation while enabling efficient reuse of motion assets across diverse character types.

With the advancement of deep learning, numerous approaches have emerged that leverage neural networks to automate skeleton-based motion retargeting. Early methods \cite{Villegas_2018_CVPR, aberman2020skeleton, 10.1145/3610548.3618206} and the recent method \cite{Hu_2024} employ RNN or Graph Neural Networks (GNNs) \cite{zhou2021graphneuralnetworksreview} to aggregate information across different skeletal structures, but these approaches require predefined network architectures tailored to specific joint configurations, limiting their generalizability across skeletons with varying numbers of joints. \cite{zhang2023skinned} utilized transformer structures to process joint features but required copying rotations directly from source to target skeletons as an initialization, restricting the application to skeletons with identical joint counts. To address topology variations, follow-up work \cite{10495176} extended this approach by proportionally copying rotations, enabling handling of different joint numbers, yet still requiring manual definition of corresponding joint chains for each skeleton pair. While \cite{martinelli2024moma} employs masked transformers to extract joint features and can process varying joint numbers, the global attention mechanism applied across all joints creates learning difficulties and hampers effective motion retargeting performance.


To address these limitations, we propose PALUM (\textbf{P}art-based \textbf{A}ttention \textbf{L}earning for \textbf{U}nified \textbf{M}otion retargeting), a novel transformer-based framework that learns skeleton-agnostic motion representations and applies them to target skeletons using their structural topology as conditioning information.
Our approach partitions skeletal joints into six semantic body parts inspired by biological musculoskeletal organization, enabling attention mechanisms to focus on joints with stronger motion correlations while reducing irrelevant cross-part interactions.
We design a spatio-temporal encoder that first applies spatial attention within each body part, incorporating joint name semantics via T5~\cite{raffel2023exploringlimitstransferlearning} embeddings and structural information through T-pose representations. This spatial processing extracts essential motion characteristics for each body part through attention pooling. A temporal transformer then processes these body-part features across time to capture motion dynamics. For retargeting, our decoder uses cross-attention to map the learned motion representation to target skeletons based on their specific topology.  To ensure robust feature learning and accurate retargeting, we introduce a cycle consistency mechanism that enforces source motion preservation through bidirectional encoding, enabling the model to maintain motion semantics while adapting to diverse skeletal topologies.
Extensive experiments demonstrate that PALUM achieves superior motion quality compared to existing methods.


In a nutshell, our main contributions are threefold:
\begin{itemize}
    \item We introduce a semantic body part grouping strategy and a spatio-temporal cross-attention mechanism combined with part-aware embeddings that effectively capture both spatial relationships between body parts and temporal motion dynamics, enabling consistent motion representation across skeletons with varying joint numbers and bone proportions;
    \item We propose a cyclic encode-decode mechanism that preserves source motion semantics while adapting to diverse skeletal structures;
    \item We demonstrate through comprehensive experiments that our approach achieves superior motion retargeting quality compared to state-of-the-art methods.
\end{itemize}



\section{Related Works}
\label{sec:related_works}

\subsection{Skeletal Variation}
Character skeletons can differ by topology, which we classify into three categories: isomorphic skeletons, homeomorphic skeletons, and non-homeomorphic skeletons. It is difficult to perform motion retargeting and motion generation with the last two skeletons. 
Graph-based approaches like Skeleton-Aware Networks \cite{aberman2020skeleton} reduce homeomorphic skeletons to a common "primal skeleton" representation using GNN.
Non-homeomorphic cases involving fundamentally different skeletons are the most challenging and historically require manual intervention, with traditional methods like \cite{:10.2312/SCA/SCA10/169-178} needing paired motion data and Creature Features \cite{Seol2013CreatureFO} requiring manual joint mapping definitions. 
Recent transformer-based approaches like \cite{martinelli2024moma} treat each joint's motion as an independent token using masked autoencoder training to learn implicit correspondences without paired data, while the AnyTop framework \cite{gat2025anytop} uses diffusion-based generation with graph-aware encoding to produce motions for arbitrary skeletal topologies. These approaches demonstrate significant progress in cross-topology motion transfer, though performance on completely unseen skeleton types may still require further investigation.

\subsection{Motion Retargeting}
Cross-skeleton motion retargeting is challenging because motions must be transferred between characters with different skeletal topologies and proportions while preserving semantic meaning. Early works like SAN~\cite{aberman2020skeleton} and SAME~\cite{10.1145/3610548.3618206} define Graph Neural Networks (GNNs) on skeleton graphs, enabling motions to be embedded into common latent spaces shared by homeomorphic skeletons. PAN~\cite{Hu_2024} follows a similar GNN approach but focuses on part-based operations. $R^2ET$\cite{zhang2023skinned} employs transformer architectures to process joint representations but maintains the constraint of identical joint counts between source and target skeletons through direct rotation transfer as initialization. MeshRet\cite{ye2024skinned} attempts to incorporate mesh information by defining a DMI field, yet its correspondence definition and reconstruction objectives still rely on rotation computations between matching joints, similarly restricting application to skeletons with identical joint numbers. Since these network structures are fundamentally defined based on joint numbers or require explicit joint matching, their generalization to characters with significantly different structural topologies remains limited.
To handle skeletons with varying joint numbers, $M-R^2ET$\cite{10495176} builds upon the transformer framework by redistributing rotation information according to bone length relationships within kinematic chains, though this still necessitates manual specification of joint chain correspondences. MoMa\cite{martinelli2024moma} applies masked transformer architectures to accommodate variable skeleton sizes, yet the global attention mechanism across all joints introduces redundant computation and learning challenges. Motion2Motion~\cite{chen2025motion2motion} proposes a lightweight, training-free retargeting method based on motion retrieval. However, this approach requires explicit joint correspondence definition between skeletons and assumes the availability of motion sequences for target skeletons, which may not always be practical.


\section{Preliminary}
\label{sec::preliminary}

\textbf{Attention pooling}. Attention pooling is a mechanism that aggregates variable-length sequences into fixed-size representations by learning to focus on the most relevant elements. It builds upon the attention mechanism popularized by transformers, but differs in its application: 
while transformers typically use attention to model relationships within or between sequences, attention pooling specifically focuses on aggregating variable-length sequences into a single fixed-size representation.

Given a sequence of tokens $\mathbf{X} = \{\mathbf{x}_1, \mathbf{x}_2, \ldots, \mathbf{x}_n\} \in \mathbb{R}^{n \times d}$, attention pooling employs the same scaled dot-product attention as transformers but with learnable queries $\mathbf{Q} = \{\mathbf{q}_1, \mathbf{q}_2, \ldots, \mathbf{q}_m\} \in \mathbb{R}^{m \times d}$, which are initialized with uniform random values and trained to extract $m$ different aspects of the input sequence.
For each query $\mathbf{q}_i$, the attention weights are computed as:
\begin{equation}
\alpha_{i,j} = \frac{\exp(\mathbf{q}_i^T \mathbf{x}_j / \sqrt{d})}{\sum_{k=1}^{n} \exp(\mathbf{q}_i^T \mathbf{x}_k / \sqrt{d})}.
\end{equation}

Then the pooled representation for query $\mathbf{q}_i$ is $\mathbf{z}_i = \sum_{j=1}^{n} \alpha_{i,j} \mathbf{x}_j$. This results in a fixed-size output $\mathbf{Z} = \{\mathbf{z}_1, \mathbf{z}_2, \ldots, \mathbf{z}_m\} \in \mathbb{R}^{m \times d}$ regardless of the input sequence length $n$. 
In our motion retargeting context, attention pooling enables us to capture common skeletal features within each body part regardless of the specific joint count, facilitating skeleton-agnostic motion representation learning.


\section{Method}

Given a source skeleton $\mathcal{S}_A$ with $N_A$ joints and a corresponding motion sequence of $T$ frames $\mathcal{M}_A = \{\mathbf{m}_A^{(1)}, \mathbf{m}_A^{(2)}, \ldots, \mathbf{m}_A^{(T)}\}$, where $\mathbf{m}_A^{(t)} = [\mathbf{r}_A^{(t)}, \mathbf{p}_A^{(t)}]$ combines the root joint position $\mathbf{r}_A^{(t)} \in \mathbb{R}^3$ and joint rotations $\mathbf{p}_A^{(t)} \in \mathbb{R}^{N_A \times 6}$ represented as 6D rotation vectors \cite{zhou2020continuityrotationrepresentationsneural} relative to their parent joints at time step $t$, our goal is to retarget this motion to a target skeleton $\mathcal{S}_B$ with $N_B$ joints, producing a semantically equivalent motion sequence $\mathcal{M}_B = \{\mathbf{m}_B^{(1)}, \mathbf{m}_B^{(2)}, \ldots, \mathbf{m}_B^{(T)}\}$, where $\mathbf{m}_B^{(t)} = [\mathbf{r}_B^{(t)}, \mathbf{p}_B^{(t)}]$ with $\mathbf{r}_B^{(t)} \in \mathbb{R}^3$ and $\mathbf{p}_B^{(t)} \in \mathbb{R}^{N_B \times 6}$.
Formally, we seek to learn a mapping function $f: (\mathcal{S}_A, \mathcal{M}_A, \mathcal{S}_B) \rightarrow \mathcal{M}_B$ that maintains motion semantics across different skeletal structures. 
The key challenge lies in the fact that $\mathcal{S}_A$ and $\mathcal{S}_B$ may have significantly different topologies, with $N_A \neq N_B$ and distinct joint connectivity patterns and different joint bone lengths. Despite these structural differences, the retargeted motion $\mathcal{M}_B$ must preserve the semantic content and naturalness of the source motion $\mathcal{M}_A$ while being anatomically plausible for the target skeleton.
This requires the model to understand the functional correspondence between body parts across skeletons and transfer motion characteristics accordingly, rather than relying on explicit joint-to-joint mappings.

In the following, we describe how we group joints in \ref{subsec:joint_group}, how we design the skeleton-agnostic feature extraction in \ref{subsec::enc}, and the skeleton-specific motion retargeting in \ref{subsec::dec}. Finally, we show the overall architecture of our training and testing pipeline in \ref{subsec::pipeline}, followed by the training objective descriptions.


\subsection{Joint Groups}
\label{subsec:joint_group}
The key insight behind our approach is that human motion exhibits inherent locality.
For instance, hand movements are largely independent of foot motions, while joints within the same limb are highly correlated due to shared neural pathways and biomechanical constraints. In mathematical terms, spatially distant joints require minimal attention, whereas proximate joints within the same kinematic chain exhibit strong interdependencies that benefit from focused attention mechanisms.

Inspired by this biological principle, we partition the skeleton into six semantic body parts: torso, left leg, right leg, left arm, right arm, and head. 
Because the torso group 
represents the primary structural axis that coordinates overall body movement and serves as the kinematic foundation for limb motions,
we design overlapping connections between groups: the left and right leg groups share the hip joint with the torso, while the left arm, right arm, and head groups share the uppermost spine joint with the torso. 
We show in the ablation study how this design helps improve the performance of motion retargeting compared with the non-shared joint groups design. 

\begin{figure*}[htb]
    \centering
    \includegraphics[width=1.\textwidth]{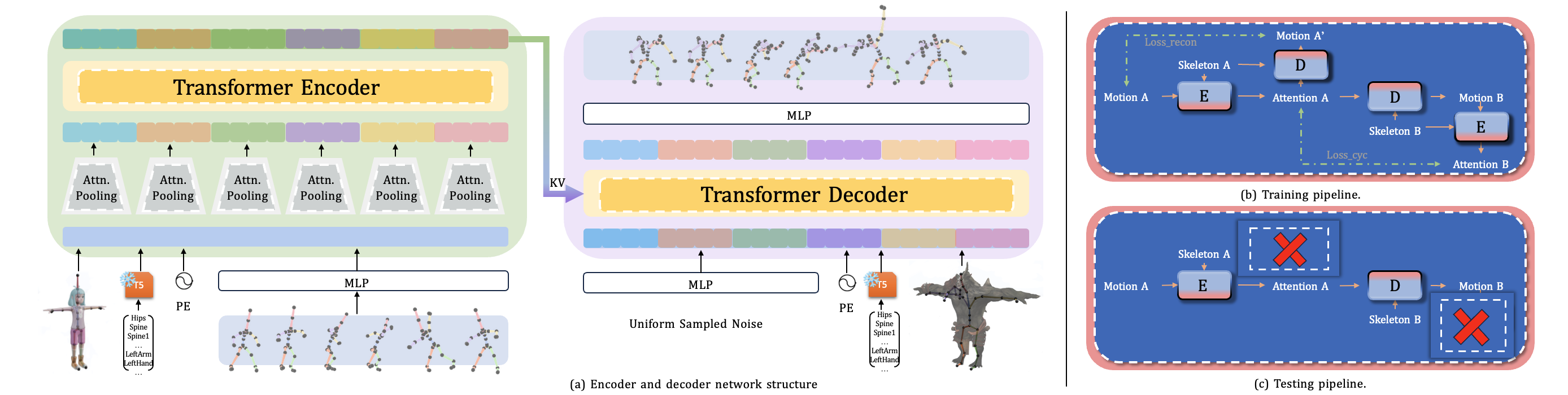}
    \caption{\textbf{Overview of our motion retargeting framework}. (a) Encoder-decoder architecture: The transformer encoder processes source motion sequences through multiple attention pooling layers to extract skeleton-agnostic motion representations. These representations are fed to an MLP that outputs key-value pairs for cross-attention in the transformer decoder. The decoder takes uniformly sampled noise and skeleton-specific embeddings to generate target motion sequences. Note that the target skeleton shown in this pipeline includes pauldron joints, demonstrating our method's capability to handle diverse skeletal topologies. (b) Training pipeline: The model is trained using reconstruction and cycle consistency objectives (additional loss terms are detailed in Section \ref{subsec::train_objective}), where Motion A from Skeleton A is retargeted to Skeleton B, then back to Skeleton A, ensuring motion preservation across different skeleton structures. (c) Testing pipeline: During inference, only the forward retargeting path is used, with the cycle consistency components (marked with red crosses) disabled, allowing direct motion transfer from the source to the target skeleton.}
    \label{fig::ae}
\end{figure*}

\subsection{Skeleton-Agnostic Feature Extraction}
\label{subsec::enc}

\subsubsection{Spatial Space Attention} 
The first part of our skeleton-agnostic feature extraction processes each semantic body part independently through a specialized spatial attention mechanism.
For a motion sequence with $T$ frames and batch size $B$, we first organize joints according to the six semantic groups defined in Section~\ref{subsec:joint_group}.
Each joint group $g_i$ ($i \in {1,2,\ldots,6}$) contains a variable number of joints $n_i$. For the root joint, we use a 12-dimensional representation including global position, 6D rotation, and velocity. For other joints, we use a 9-dimensional representation consisting of root-relative position and 6D rotation. These features are processed through an MLP to obtain a unified feature dimension $D$. To enable efficient batch processing, we pad each group to its maximum joint count $\max(n_i)$, resulting in group representations $\mathbf{X}^{(i)}$ of size $\mathbb{R}^{T \times B \times \max(n_i) \times D}$.
For each joint within a group, we incorporate three types of embeddings to provide rich contextual information.
\begin{itemize}
    \item \textbf{Positional encoding}: sinusoidal positional encodings $\mathbf{E}_{PE}^{(i)} \in \mathbb{R}^{\max(n_i) \times D}$ are applied to distinguish joint positions within each group.
    \item \textbf{Joint name embeddings}: joint names extracted from the original BVH motion files are encoded using the T5 text encoder~\cite{raffel2023exploringlimitstransferlearning}. The resulting joint name embeddings $\mathbf{E}_{name}^{(i)} \in \mathbb{R}^{\max(n_i) \times D}$ capture semantic relationships between anatomically similar joints.
    \item \textbf{T-pose embeddings}: T-pose embeddings $\mathbf{E}_{tpose}^{(i)} \in \mathbb{R}^{\max(n_i) \times D}$ are derived from joint root-relative positions in the T-pose configuration, providing structural information about the skeleton topology. Unlike previous methods \cite{Guo_2022_CVPR, gat2025anytop} that incorporate joint rotations, velocities, and contact labels, only positional information is used to avoid rotation inconsistencies across different coordinate system conventions that can hinder training convergence and generalization.
\end{itemize}
The T-pose representation is first padded to the max joint number and processed through an MLP to get the T-pose feature $E_{T-pose}^{i} \in \mathbb{R}^{\max(n_i)} \times D$, then added to every frame of the motion embedding, ensuring that skeletal structural information is consistently available during feature learning and preventing topology information loss.

After the enhancement, we get 
\begin{equation}
    \mathbf{X}_{enhanced}^{(i)} = \mathbf{X}^{(i)} + \mathbf{E}_{PE}^{(i)} + \mathbf{E}_{name}^{(i)} + \mathbf{E}_{T-pose}^{(i)}.
\end{equation}
We then apply attention pooling along the joint axis for each group: $\mathbf{Z}^{(i)} = \text{AttentionPooling}(\mathbf{X}_{enhanced}^{(i)}, \mathbf{Q}^{(i)})$,
where $\mathbf{Q}^{(i)} \in \mathbb{R}^{m \times D}$ are the learnable query vectors for group $i$, and $\mathbf{Z}^{(i)} \in \mathbb{R}^{T \times B \times m \times D}$ represents the pooled features.
Finally, we concatenate the outputs from all six groups to obtain a unified representation $\mathbf{Z} \in \mathbb{R}^{T \times B \times M \times D}$, where $M = m \times 6$, to capture the essential motion characteristics across all semantic body parts.

\subsubsection{Temporal Space Attention}
Following the spatial attention pooling, we process the temporal dynamics through our main transformer encoder blocks. The spatial feature $\mathbf{Z}$ from the previous stage serves as input to this temporal processing module. 
Multi-head attention is applied across all $M$ tokens to capture both temporal dependencies within the motion sequence and cross-part interactions between different body regions simultaneously.
The output of the temporal encoder is a refined representation $\mathbf{H} \in \mathbb{R}^{T \times B \times M \times D}$ that encodes both spatial motion characteristics and their temporal evolution, serving as a comprehensive motion representation for the subsequent retargeting process.
The structure of the motion feature extractor is shown on the left side of Figure. \ref{fig::ae}.(a).

\subsection{Skeleton-Specific Motion Retargeting}
\label{subsec::dec}


\subsubsection{Retargeting Input Construction}
The retargeting input is initialized with random uniform noise $\mathbf{Y}_{init} \in \mathbb{R}^{T \times B \times N_B \times d}$, where $N_B$ is the number of joints in the target skeleton and $d$ represents the joint feature dimension (12 for root joint, 9 for others). The noise is then organized into the six semantic groups and padded to each group's maximum joint count $\max(n_i)$ for parallel processing. Each group's noise is processed through an MLP to obtain the unified feature dimension $D$, resulting in decoder input representations of size $\mathbb{R}^{T \times B \times \max(n_i) \times D}$ for each group.

To provide the retargeting module with essential information about the target skeleton structure and pose, we also incorporate three types of embeddings:
\begin{itemize}
    \item \textbf{Positional embeddings}: target-specific sinusoidal positional encodings $\mathbf{E}_{PE}^{(i)} \in \mathbb{R}^{N_B \times D}$ distinguish joint positions within the target skeleton hierarchy.
    \item \textbf{Target joint name embeddings}: joint names from the target skeleton are encoded using the same T5 text encoder, producing embeddings $\mathbf{E}_{name}^{(i)} \in \mathbb{R}^{N_B \times D}$ that capture semantic correspondences with source joints.
    \item \textbf{T-pose embeddings}: the target skeleton's T-pose configuration $\mathbf{E}_{T-pose}^{i} \in \mathbb{R}^{N_B \times D}$ provides structural constraints and default joint relationships specific to the target anatomy.
\end{itemize}
The enhanced decoder input is formulated as 
\begin{equation}
    \mathbf{Y}_{input}^{i} = \mathbf{Y}_{init} + \mathbf{E}_{PE}^{(t)} + \mathbf{E}_{name}^{(t)} + \mathbf{E}_{T-pose}^{i}.
\end{equation}

\subsubsection{Retargeting Module Architecture}
The retargeting module employs a standard transformer decoder architecture with both self-attention and cross-attention mechanisms.
The cross-attention mechanism uses the feature extraction output $\mathbf{H}$ as both keys and values, enabling the decoder to attend to relevant motion patterns from the source skeleton while generating motion for the target skeleton.
The retargeting module output $\mathbf{Y}_{final} \in \mathbb{R}^{T \times B \times \sum\max(n_i) \times D}$ is first processed by masking out padding tokens to obtain the actual target joint features of size $\mathbb{R}^{T \times B \times N_B \times D}$. We then extract the predicted joint rotations and apply forward kinematics (FK) with the predefined topology of skeleton $\mathcal{S}_B$ to compute global joint positions. The root movements are obtained by normalizing with respect to the T-pose root height of the source skeleton $\mathcal{S}_A$ and then de-normalizing using the T-pose root height of the target skeleton $\mathcal{S}_B$.
The structure of the motion decoder is shown on the right side of Figure. \ref{fig::ae}.(a).

\subsection{Overall Architecture}
\label{subsec::pipeline}

Figure~\ref{fig::ae} illustrates the complete pipeline of our cross-skeleton motion retargeting framework when training and testing separately.

For training, given a source motion sequence $\mathcal{M}_A$ performed by skeleton $\mathcal{S}_A$, and a target skeleton $\mathcal{S}_B$, the feature extraction module $E$ takes the source motion and skeleton as input and produces a semantic motion representation: $\mathbf{H_A} = E(\mathcal{M}_A, \mathcal{S}_A) \in \mathbb{R}^{T \times B \times M \times D}$ captures the essential motion characteristics through our spatial-temporal attention mechanism.
Then the retargeting module $D$ utilizes both the motion representation $\mathbf{H_A}$ and the target skeleton structure $\mathcal{S}_B$ to generate the retargeted motion $\mathcal{M}_B = D(\mathbf{H_A}, \mathcal{S}_B)$, where $\mathcal{M}_B$ represents the motion sequence adapted for skeleton $\mathcal{S}_B$. The predicted motion $\mathcal{M}_B$ and the target skeleton $\mathcal{S}_B$ are sent into the same feature extraction module $E$ to get the semantic motion representation $\mathbf{H_B} = E(\mathcal{M}_B, \mathcal{S}_B) \in \mathbb{R}^{T \times B \times M \times D}$ that captures the essential motion characteristics for motion $\mathcal{M}_B$ and skeleton $\mathcal{S}_B$. Besides that, $\mathbf{H_A}$ and $\mathcal{S}_A$ are also sent into the same retargeting module $D$ to get predicted motion $\mathcal{M}_{A'} = D(\mathbf{H_A}, \mathcal{S}_A)$.

\subsubsection{Motion Reconstruction Training}
To enable the motion feature extractor to effectively capture spatial-temporal information from motion sequences conditioned on specific skeleton structures, and to ensure the retargeting module can accurately generate motion based on attention mechanisms and skeleton constraints, we employ motion reconstruction training. This self-supervised approach trains the model to reconstruct the original motion from its learned representation.
Specifically, for a given motion-skeleton pair $(\mathcal{M}_A, \mathcal{S}_A)$, we first extract the motion features to obtain the semantic representation $\mathbf{H_A}$, then use the same retargeting module to reconstruct the original motion $\mathcal{M}_{A'} = D(\mathbf{H_A}, \mathcal{S}_A)$.
The reconstruction loss is formulated as a Mean Squared Error (MSE) between the original motion and the reconstructed motion
\begin{equation}
\mathcal{L}_{rec} = \text{MSE}(\mathcal{M}_A, \mathcal{M}_{A'}) = \frac{1}{N} \sum_{i=1}^{N} ||\mathcal{M}_A^{(i)} - \mathcal{M}_{A'}^{(i)}||_2^2,
\end{equation}
where $N$ represents the total number of joint-frame pairs in the motion sequence. This reconstruction objective ensures that the feature extraction and retargeting modules learn to preserve essential motion information while being conditioned on the specific skeletal structure, providing a strong foundation for cross-skeleton motion retargeting."

\subsubsection{Cycle Consistency Training}
While motion reconstruction training ensures the feature extraction and retargeting modules preserve motion information for a given skeleton, it does not guarantee that the learned representations are skeleton-agnostic. To address this limitation, we introduce cycle consistency training that enforces the feature extraction module to learn motion features independent of specific skeleton structures.
The key insight behind cycle consistency training is that if the feature extraction module truly captures skeleton-invariant motion semantics, then the attention representations $\mathbf{H_A}$ and $\mathbf{H_B}$ derived from the same underlying motion should be similar, regardless of the different skeletons on which they are conditioned. By enforcing this consistency, we encourage the feature extraction module to focus on the intrinsic motion characteristics rather than skeleton-specific details.
Specifically, given the attention representations $\mathbf{H_A} = E(\mathcal{M}_A, \mathcal{S}_A)$ from the source motion-skeleton pair and $\mathbf{H_B} = E(\mathcal{M}_B, \mathcal{S}B)$ from the retargeted motion-skeleton pair, we formulate the cycle consistency loss as
\begin{equation}
\mathcal{L}{cyc} = \frac{1}{T \times B \times M \times D} \sum |\mathbf{H_A} - \mathbf{H_B}|_2^2,
\end{equation}
where the summation is over all temporal frames $t$, batch samples $b$, semantic joints $j$, and feature dimensions $d$.

\subsubsection{Target Positions for Motion B} 
A fundamental question in motion retargeting is determining appropriate target positions for end-effector joints in skeleton B. We consider two positioning strategies: relative positioning based on joint chain length ratios, and absolute positioning that preserves world coordinates. The optimal choice depends on the motion context. For example, for dance movements, relative positioning better preserves the motion's aesthetic proportions, while for object interaction tasks like door opening, absolute positioning ensures proper contact with environmental constraints (e.g., door handles).
We train three variants to evaluate these approaches: (1) end-effector positions computed using relative joint chain length ratios between source and target skeletons, (2) absolute source positions for arm end-effectors, and (3) unconstrained end-effector positions. Detailed implementation and comparative results are provided in the supplementary material.
All qualitative and quantitative results presented in the main paper use approach (3) with unconstrained end-effector positions.

\subsubsection{Inference Pipeline}
For testing, our framework performs motion retargeting in a straightforward forward pass, as illustrated in Figure~\ref{fig::ae}. (c). Given a source motion sequence $\mathcal{M}_A$ and target skeleton $\mathcal{S}_B$, the trained feature extraction module $E$ processes the source motion and its corresponding skeleton to extract the skeleton-agnostic motion representation:
$\mathbf{H_A} = E(\mathcal{M}_A, \mathcal{S}_A)$.
Then, the trained retargeting module $D$ takes this motion representation along with the target skeleton structure to generate the retargeted motion $\mathcal{M}_B = D(\mathbf{H_A}, \mathcal{S}_B)$.

\subsection{Training objective.}
\label{subsec::train_objective}
Our complete training objective combines the reconstruction loss with additional regularization terms to ensure motion quality:

\begin{equation}
\mathcal{L}_{total} = \mathcal{L}_{rec} + \lambda_{cyc}\mathcal{L}_{cyc} +
\lambda_{root}\mathcal{L}_{root},
\end{equation}


where $\mathcal{L}_{rec}$ and $\mathcal{L}_{cyc}$ are the reconstruction and cycle consistency losses described above, and the additional terms provide crucial constraints for realistic motion generation.

\textbf{Root stability loss:} The root stability loss $\mathcal{L}_{root}$ ensures that the global positioning and orientation of the character remain consistent during reconstruction. This loss is computed as the mean squared error between the root position and rotation of the original motion $\mathcal{M}_A$ and the reconstructed motion $\mathcal{M}_{A'}$: 
\begin{equation}
    \mathcal{L}_{root} = \text{MSE}(\mathbf{p}_{root}^A, \mathbf{p}_{root}^{A'}) + \text{MSE}(\mathbf{r}_{root}^A, \mathbf{r}_{root}^{A'}),
\end{equation}
where $\mathbf{p}_{root}$ and $\mathbf{r}_{root}$ represent the root position and rotation respectively. This constraint is particularly important for maintaining character stability and preventing unrealistic drifting or spinning behaviors.

The hyperparameters $\lambda_{cyc}$ and $\lambda_{root}$ control the relative importance of each loss component. We set $\lambda_{cyc} = 20$, $\lambda_{root} = 7$.

\section{Experiments}

\textbf{Implementation details.}
Our method is implemented in PyTorch and trained on NVIDIA A100 GPUs. We use the AdamW optimizer with a starting learning rate of $1e-4$, momentum parameters $\beta_1 = 0.9$ and $\beta_2 = 0.99$, and weight decay of $0.999$. The training is conducted with a batch size of 16 and a motion window length of 64 frames.

\begin{figure}[htb]
    \centering
    \includegraphics[width=0.45\textwidth]{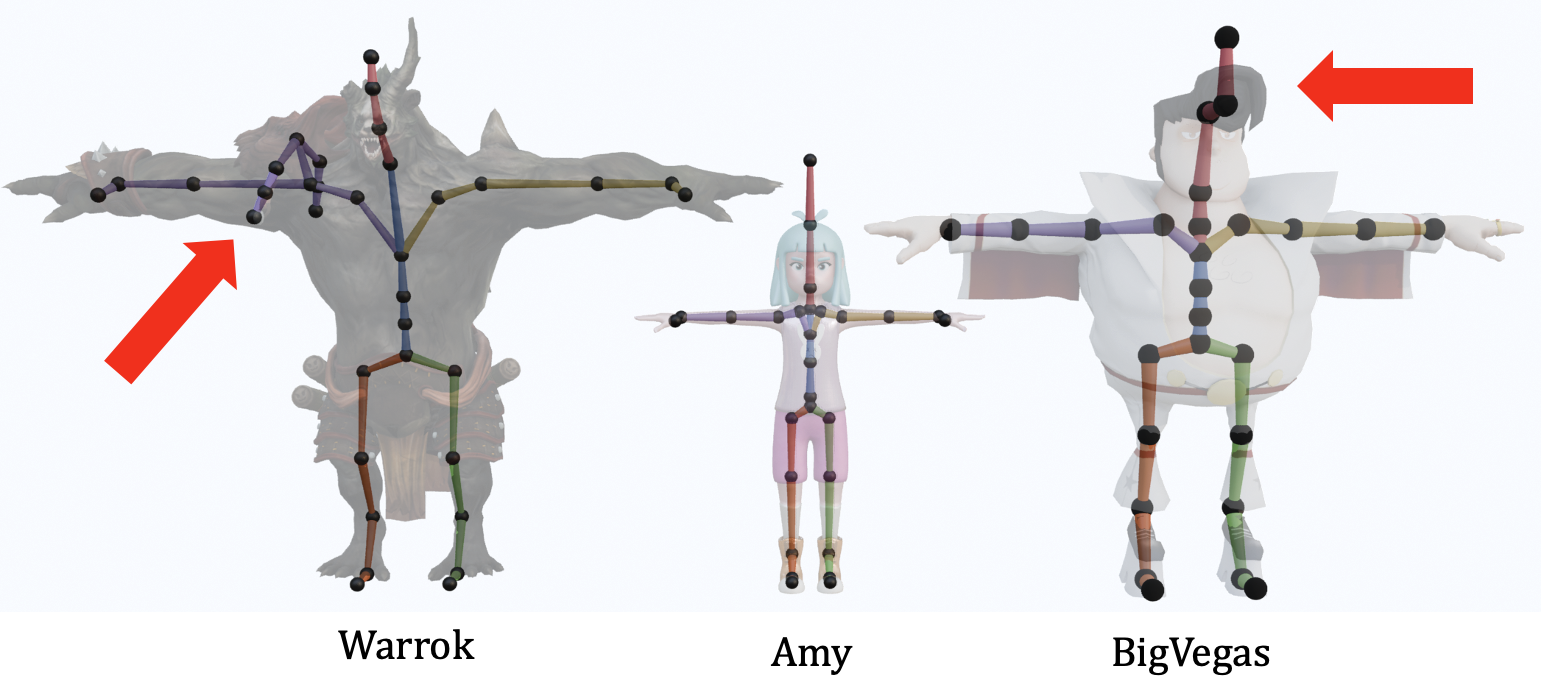}
    \caption{T-pose examples after our joint elimination strategy. (1) Warrok: the pauldron joints are named as "RightArmourx" (x=1,2,3,4,5) in the BVH file, so they match our "RightArm" joint name selection and they are preserved. (2) BigVegas: the hair joints are named as "HeadTop\_Endx" (x=1,2), which match our "Head" joint name selection, so they are preserved.}
    \label{fig::3_tpos}
\end{figure}

\textbf{Datasets.}
We evaluate our method on the Mixamo dataset~\cite{mixamo}. We collect 12 characters for training and 7 characters for testing, which in total results in 3512 motion sequences at 30 frames per second.
Following previous methods~\cite{martinelli2024moma, 10.1145/3610548.3618206, zhang2023skinned, aberman2020skeleton}, we eliminate finger joints from all skeletons to focus on body motion retargeting. 
However, unlike approaches that select joints by specific joint names, we remove joints by filtering out those containing finger-related identifiers (e.g., "finger", "thumb", "index") and other joints whose names do not match the selections. This deletion-based approach results in skeletons with varying numbers of remaining joints, as shown in Figure. \ref{fig::3_tpos}. Besides that, we also follow SAN \cite{aberman2020skeleton} to split joints to get more varied skeleton structures with different joint numbers. 
To comprehensively evaluate our method's generalization capability, we construct four evaluation splits based on the visibility of characters and motions during training: We have unseen character (uc), unseen motion (um), seen character (sc), and seen motion (sm), resulting in four evaluation scenarios: sc+sm, sc+um, uc+sm, uc+um. 

\textbf{Baselines.} We compare our method with three recent state-of-the-art skeleton-based approaches for motion retargeting: R$^2$ET \cite{zhang2023skinned}, PAN \cite{Hu_2024} and MoMa \cite{martinelli2024moma}. 
R$^2$ET and MoMa are shape-aware methods that employs separate networks for skeletal structure and character geometry. Following our problem setting, we train and evaluate only the skeleton network components. Since R$^2$ET is limited to intra-structural motion retargeting, we compare against it only on the intra-structural scenarios.  
For fair evaluation, we train all the three baselines on our training dataset and evaluate them on our test set using identical experimental protocols. The training dataset contains no paired motion sequences across different characters. 
Note that we train and test baseline methods using their original joint selection strategies, while our method is trained and tested using our joint elimination strategy, which presents a more challenging evaluation setting for our approach. For intra-structural experiments, our method retains auxiliary joints such as pauldron joints, while cross-structural data follows the splitting protocol established by SAN.
We evaluate the retargeted motions by comparing the global joint positions of the generated sequences against the ground truth motions for the target skeleton. The evaluation metric is Mean Squared Error (MSE) normalized by each character's height.

\subsection{Quantitative Results}
We compare our method with the three baselines in the four evaluation scenarios defined above. The quantitative results are shown in Table \ref{tab::quan}. Our method demonstrates superior performance across all test conditions, achieving the lowest MSE in both intra-structural and cross-structural motion retargeting tasks. 
Notably, our approach shows particularly strong improvements in the challenging cross-structural scenarios, where the skeletal differences are most pronounced. 

\begin{table*}[htbp]
\centering
\setlength{\tabcolsep}{4pt}
\small
\begin{tabular}{l|c|c|c|c|c|c|c|c|c|c}
\hline
 & \multicolumn{5}{c|}{Intra-Structural} & \multicolumn{5}{c}{Cross-Structural} \\
\hline
 & sc+sm & sc+um & uc+sm & uc+um & mean & sc+sm & sc+um & uc+sm & uc+um & mean \\
\hline
R$^2$ET & 0.00424 & 0.00637 & 0.00637 & 0.00249 & 0.00487 & - & - & - & - & - \\
PAN & 0.00629 & 0.00852 & 0.01030 & 0.01079 & 0.00898 & 0.01294 & 0.01412 & 0.01378& 0.01660 & 0.01436 \\
MoMa &  0.01980 &  0.00711 & 0.01517 & 0.00193 &  0.01100 & 0.03687 & 0.01857 & 0.02833 & 0.00983 & 0.02340 \\
\hline
Ours & \textbf{0.00214} & \textbf{0.00343} & \textbf{0.00341} & \textbf{0.00188} & \textbf{0.00272} & \textbf{0.00225} & \textbf{0.00755} & \textbf{0.004920} & \textbf{0.00795} & \textbf{0.00567} \\
\end{tabular}
\caption{Comparison with state-of-the-art methods on intra-structure and cross-structure motion retargeting.}
\label{tab::quan}
\end{table*}

\subsection{Qualitative Results}
\begin{figure*}[htb]
    \centering
    \includegraphics[width=0.95\textwidth]{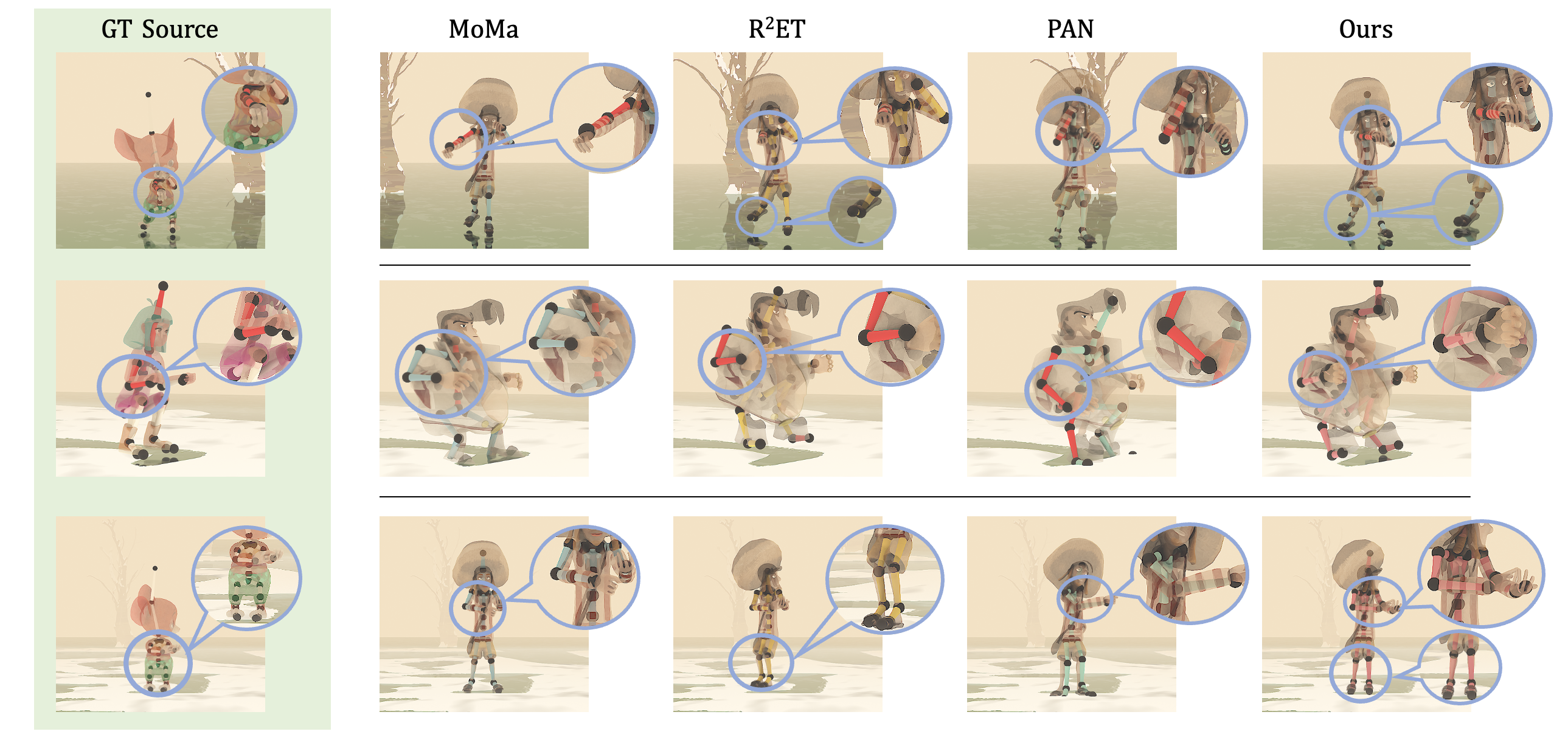}
    \caption{Qualitative results of our method and the baselines. Our method preserves the natural motion dynamics and joint relationships.}
    \label{fig::qual}
\end{figure*}

Figure \ref{fig::qual} presents qualitative comparisons of motion retargeting results across different methods. Our approach produces more natural and semantically consistent motions compared to the baselines. The difference in joint count results from our joint deletion policy, in contrast to their joint selection approach.
The results show that $R^2ET$ preserves overall motion semantics at first glance because their method initially copies rotations from the source skeleton to the target skeleton through joint mapping, followed by optimization. However, their proposed Distance Matrix (DM) constraint is overly restrictive, producing conservative retargeting results, particularly for end-effector joints, which fail to preserve the original motion style. For example, in the first row of Figure \ref{fig::qual}, the source "Cross Punch" motion exhibits "wilder" movements than its generated output.
For PAN, since it processes each body part independently, it often achieves good results for individual body parts while producing artifacts in others, leading to inconsistent overall motion semantics and reduced global coherence. MoMa, which neither copies initial rotations nor processes joints by semantic body parts, instead processes all joints collectively through a transformer with masking. This approach results in globally inaccurate motion trends, as the method struggles to capture the hierarchical relationships and semantic structure inherent in skeletal motion.
In contrast, our method effectively balances motion preservation and adaptation by leveraging semantic body part grouping with attention mechanisms, producing retargeted motions that maintain both the dynamic characteristics and stylistic nuances of the source motion while ensuring natural coordination across all body parts.
Our method can also retarget motions to SMPL \cite{SMPL:2015} and MetaHuman \cite{metahuman}, as shown in Figure \ref{fig::limitation}. More results are in the Supplementary.

\subsection{Ablation Study}
\begin{figure}[htb]
    \centering
    \includegraphics[width=0.5\textwidth]{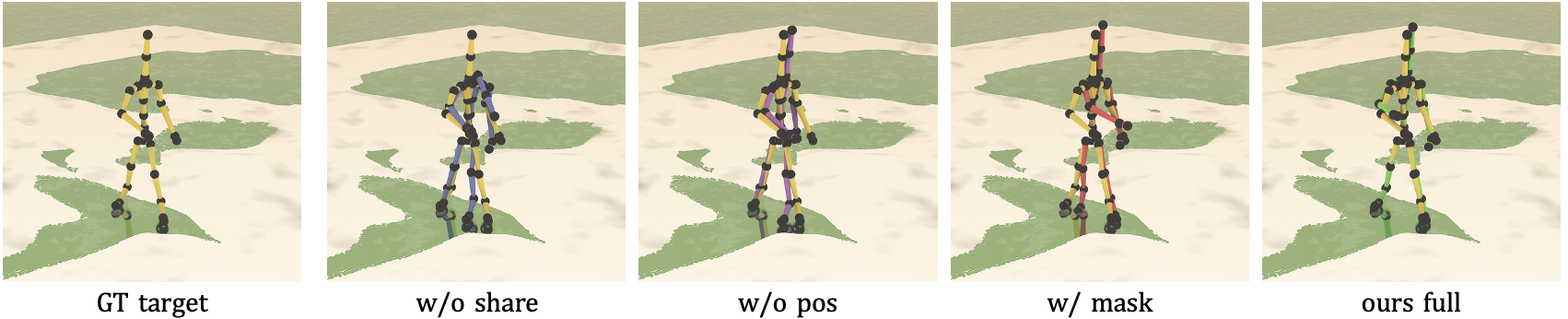}
    \caption{Qualitative results of ablation studies demonstrating the impact of key design components. We show overlaid predictions with the GT target. Without shared joints, body parts move independently, causing unnatural artifacts such as extreme head tilting. Excluding positional information reduces end-effector accuracy, while joint masking disrupts hierarchical skeletal relationships, leading to degraded motion quality.}
    \label{fig::abla}
\end{figure}

To validate the effectiveness of each component in our proposed method and understand their individual contributions to motion retargeting performance, we conduct comprehensive ablation studies on key design choices.
We systematically analyze the following components: 
\begin{itemize}
    \item \textbf{Group joint policy}: We examine whether allowing shared joints between different body parts improves the grouping strategy compared to strictly disjoint body part assignments.
    \item \textbf{Joint masking}: Inspired by MoMa's masking strategy, we evaluate the impact of randomly masking joints to zero before attention pooling to assess whether this approach improves performance in our method.
    \item \textbf{Motion representation}: We compare using the full motion representation, including both joint positions and rotations, versus without positional information.
\end{itemize}
Results shown in Table.\ref{tab::ablation} and Figure.\ref{fig::abla} indicates that:
(1) Allowing shared joints between body parts significantly improves performance across all metrics. Shared joints provide vital connections between different body parts. Without shared connections, body parts can move independently (in Figure.\ref{fig::abla}, the head bends down with an extremely large angle), leading to unnatural motion artifacts and reduced overall quality.
(2) Without positional information in the input features, the retargeted motions exhibit reduced accuracy at end-effector joints.
(3) The random joint masking strategy degrades our method's performance. This is because our input uses relative rotations with respect to parent joints, and randomly masking joints disrupts the network's understanding of hierarchical skeletal relationships within the motion. Additionally, masking reduces the information available during attention pooling, resulting in latent features $\mathbf{H}$ that only partially capture the common motion characteristics shared across different skeleton topologies.

\begin{table}[htbp]
\centering
    \centering
    \setlength{\tabcolsep}{2pt}
    \begin{tabular}{l|c|c}
    \hline
     & Intra-Structural & Cross-Structural \\
    \hline
    w/o share & 0.00377 & 0.00653 \\
    w/o pos & 0.00350 & 0.00660 \\
    w/ mask & 0.00343 & 0.00693 \\
    \hline
    Our's Full & \textbf{0.00272} & \textbf{0.00567} \\
    \end{tabular}
    \caption{Quantitative ablation study results evaluating key design components of our method on both intra-structure and cross-structure motion retargeting performance.}
    \label{tab::ablation}
\end{table}



\section{Limitations}
\begin{figure}
    \centering
    \includegraphics[width=0.45\textwidth]{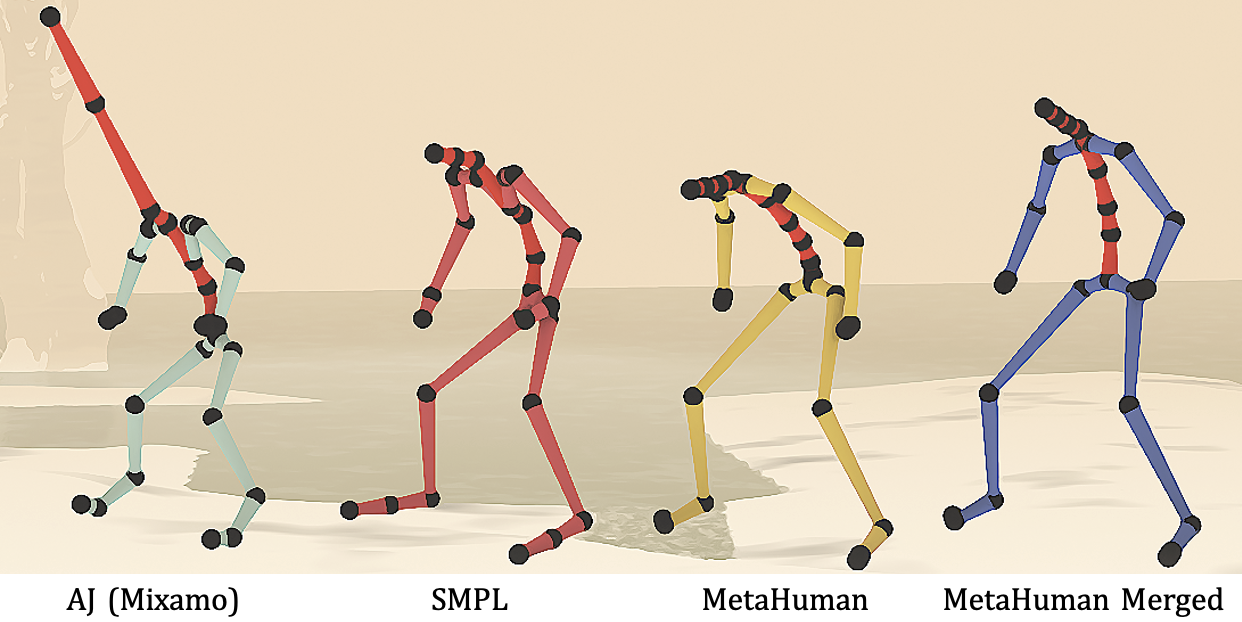}
    \caption{Mixamo motion retargeted to other skeletons. Source motion from the AJ skeleton is retargeted to SMPL, MetaHuman, and MetaHuman Merged.}
    \label{fig::limitation}
\end{figure}

Our T5-based joint name encoding creates semantic embeddings that handle naming variations (e.g., $"left\_shoulder"$, $"L\_shoulder"$, $"LeftShoulder"$) but has inherent limitations when skeletons have different joint counts along kinematic chains. The model predicts similar rotations for similarly named joints regardless of kinematic chain length. When retargeting from Mixamo's 4-joint spine to MetaHuman's 6-joint spine \cite{metahuman}, each joint contributes similar bending, causing excessive overall curvature (Figure \ref{fig::limitation}. SMPL and MetaHuman Merged exhibit similar bending curvature, while MetaHuman shows excessive spinal flexion due to its 6-joint chain structure). We currently mitigate this by merging intermediate joints during retargeting (MetaHuman Merged is created by merging spine joints [spine01, spine02, spine03] into a single spine03 joint, resulting in a 4-joint spine chain) and then recovering by setting their rotations to zero. Future work will explore more generalizable features like bone length ratios to better capture skeletal structure beyond naming semantics.

\section{Conclusion}

In this paper, we present a novel skeleton-agnostic motion retargeting method that transfers motions across characters with diverse skeletal structures. Our approach leverages spatial attention pooling with semantic body part grouping to extract common motion features across different skeleton topologies, while a temporal encoder captures temporal motion dynamics. The cycle consistency mechanism ensures robust feature learning and accurate motion preservation.
Future work includes exploring more generalizable motion representations and extending our approach to diverse skeleton structures such as quadrupeds and other animal forms. We also plan to investigate applying this framework to related tasks, including motion generation, motion editing, and motion-in-between.
We hope this work contributes to advancing cross-character motion transfer techniques and may be beneficial for applications where diverse character motions are required.
{
    \small
    \bibliographystyle{ieeenat_fullname}
    \bibliography{main}
}


\end{document}